\documentclass[11pt,a4paper]{article}
\usepackage[T1]{fontenc}
\usepackage{wrapfig}
\usepackage{url}
\usepackage{adjustbox}
\usepackage{multicol}
\usepackage{capt-of}
\usepackage{enumitem}
\usepackage{calc}
\usepackage{amssymb}
\usepackage{amsmath}
\usepackage{amsthm}
\usepackage{multicol}
\usepackage{cite}
\usepackage{graphicx}
\usepackage{enumerate}
\usepackage[svgnames]{xcolor}
\usepackage{breqn}
\usepackage{framed}
\usepackage[pdftex]{hyperref}

\usepackage[left=1.in, right=1.in, top=1in, bottom=1in]{geometry}

\usepackage{fancyhdr}
\theoremstyle{definition}

\theoremstyle{remark}

\numberwithin{equation}{section}

\usepackage{array}
\makeatletter
\newcommand{\svhline}{%
	\noalign{\ifnum0=`}\fi
	\hrule height 2\arrayrulewidth
	\futurelet\reserved@a\@xhline
}
\makeatother

\providecommand{\keywords}[1]{\noindent\small\textbf{Keywords:} #1}
\providecommand{\msc}[1]{\noindent\small\textbf{MSC 2020:} #1}

\title{\large\bfseries Trajectory-Based Bayesian Discovery of a Host-Pathogen Model from Quasi-Steady-State Observations}

\author{
	\small Reymart Salcedo Lagunero\thanks{
		\texttt{reymartl@ruc.dk} |
		Department of Science and Environment, Roskilde University,
		Universitetsvej 1, P.O. Box 260, DK-4000 Roskilde, Denmark.
	}
	\and
	\small Andreas Langer\thanks{
		\texttt{andreas.langer@math.lth.se} |
		Centre for Mathematical Sciences, Lund University,
		Box 118, 221\,00 Lund, Sweden.
	}
	\and
	\small Cordula Reisch\thanks{
		\texttt{cordular@ruc.dk} |
		Department of Science and Environment, Roskilde University,
		Universitetsvej 1, P.O. Box 260, DK-4000 Roskilde, Denmark.
	}
}

\date{\small 11. November 2026}

\begin{document}
	
	\maketitle
	
	\begin{abstract}
		System identification of nonlinear dynamical systems becomes challenging when the measured data come from a nearly settled state. We study this situation in the case of a bistable host-pathogen interaction model with a cubic term, whose seven true mechanisms are hidden among twenty candidate terms. The data are from one noise-free trajectory on a quasi-steady-state observed window and the initial state. In this situation, classical sparse identification approaches fail to identify the correct mechanisms as the candidate library becomes ill-conditioned along the quasi-steady-state trajectory. To address this, we propose an alternative approach in which sparse Bayesian learning and $\ell_{0}$ iterative hard thresholding with pursuit provide a good start, and a two-swap climb on the Bayesian evidence of the candidate ODE solutions recovers the true model. Our results show successful model recovery, i.e., correct identification of its support and accurate estimation of its coefficients, as validated by exhaustive scoring.
	\end{abstract}
	
	
	\msc{93B30; 62F15; 34A55; 68T07; 92B20.}
	
	\section{Problem description}\label{sec:1}
	
	Many mathematical models in the natural sciences need to be built from few measurements, often collected after the initial transient dynamics have passed: a chronic liver infection may be observed only after it has entered a persistent phase, a population may be counted close to its carrying capacity \cite{reisch_modeling_2019,reisch_model_2025}. We call this nearly settled phase the \textit{quasi-steady-state regime} and ask whether the mechanisms driving such a system into this regime can be recovered from data that carry almost no dynamics.
	
	In \cite{reisch_model_2025}, the authors investigated this problem using a quadratic competition model of two species. Starting from an enlarged set of candidate mechanisms, they attempted to recover the governing equations using observations restricted to the quasi-steady-state regime, together with the initial state. Their iterative coefficient-pruning algorithm retained one spurious mechanism and produced coefficient estimates that deviated from the ground truth. The sparse identification of nonlinear dynamics (SINDy) method \cite{Brunton_Proctor_Kutz_2016} retained even more spurious mechanisms, with some coefficients orders of magnitude larger than their true values.
	
	In this work, we present a trajectory-based Bayesian approach for identifying the underlying mechanisms of a dynamical system and estimating their corresponding coefficients from quasi-steady-state observations. We demonstrate that, for the numerical setting considered, the approach can recover the correct mechanisms and estimate their coefficients with high accuracy.
	
	{\bfseries Model:}
	For $t>0$, we consider the host-pathogen interaction model
	\begin{equation}\label{eqn:cubicSystem}
		\dot{u} = - a_{1} u^{3} + a_{2} u^{2} - a_{3} u  - a_{4} u v, \qquad \dot{v} = b_{1} u - b_{2} v + b_{3} u v,
	\end{equation}
	with $u(0) = u_{0},  v(0) = v_{0}$ and with parameters $a_{i}$ and $b_{j}$ assumed to be positive.
	The model describes the evolution of a host population $u := u(t) \in \mathbb{R}$ with a strong Allee effect on its logistic growth and a predator-prey-like decay. The growth of the pathogen population $v := v(t) \in \mathbb{R}$ is controlled through $u$ and a predator-prey-like growth term, and a natural decay limits its growth. 
	For $t\geq 0$, we collect the two states in $x(t) := (u(t), v(t)) \in \mathbb{R}^{2}$, write $\dot{x}(t) = (\dot{u}(t), \dot{v}(t))$ for its time derivative, and $x_{0} := (u_{0}, v_{0})$ for the initial state. Here and in the following, unless otherwise stated, a tuple such as $(u,v)$ denotes a column vector, and a tuple of column vectors denotes the column that stacks them.
	
	\textbf{Equilibria:} We use $(a_1, a_2, a_3, a_4, b_1, b_2, b_3) = (1.0, 1.1, 0.2, 0.2, 0.5, 1.0, 0.5)$ as ground truth parameters. System \eqref{eqn:cubicSystem} has three equilibria on the nonnegative quadrant: the extinction state $(0, 0)$, a stable node; the shifted Allee threshold $(0.2875,0.1679)$, a saddle point; and the coexistence state $ x^{*} := (0.6407,0.4714)$, a stable spiral. A fourth equilibrium lies at $v < 0$ and plays no biological role. Hence, the system is bistable in this parameter setting and the initial state decides the long-term behaviour of solutions.
	
	Data taken near the coexistence state constrain the model only there. As Section~\ref{sec:3} shows, a model with wrong mechanisms can also accurately fit the data while predicting different behaviour away from them, and so the mechanisms should be identified to support reliable predictions.
	
	{\bfseries Data:} One noise-free trajectory of \eqref{eqn:cubicSystem} from the initial state $x_{0} = (1.0, 0.5)$ (LSODA \cite{Petzold_1983}, relative and absolute tolerances $10^{-10}$ and $10^{-12}$, resp.) is observed at $N_{d}=100$ evenly spaced times on the window $[10, 20]$, $x_{0}$ is known. For $t_{i} \in [10,20]$, $i=1,\ldots, N_{d}$, we write $y_{i} := x(t_{i}) \in \mathbb{R}^{2}$ and $y := (y_{1}, \ldots, y_{N_{d}}) \in \mathbb{R}^{n}$ with $n = 2 N_{d} = 200$ for the observations and their stacked vector respectively. Note that $x_{0}$ is not part of $y$. solution
	
	\subsection{Computational recovery of the system}\label{subsec:1:2}
	
	\textbf{Library:} We collect all $M=10$ monomials of degree at most three in the row vector $\Theta(x):=(\Theta_{1}(x),\ldots,\Theta_{M}(x))=(1,u,v,u^{2},uv,v^{2},u^{3},u^{2}v,uv^{2},v^{3})\in\mathbb{R}^{1\times M}$. A candidate model has a coefficient matrix $\Xi=(\xi_{u},\xi_{v})\in\mathbb{R}^{M\times2}$ with one column per equation, and reads
	\begin{equation}\label{eqn:candidateModel}
		\dot{u}=\Theta(x)\xi_{u},\quad \dot{v}=\Theta(x)\xi_{v},\quad\text{that is,}\quad \dot{x}=f(x;\Xi):=\Xi^{\top}\Theta(x)^{\top}\in\mathbb{R}^{2}.
	\end{equation}
	An entry $\Xi_{mj}$, $j\in\{u,v\}$, is the coefficient of $\Theta_{m}$ in the equation of $j$; we call it a \textit{mechanism}, e.g., $uv\to\dot{u}$. There are $2M=20$ mechanisms, and \eqref{eqn:cubicSystem} has seven.
	
	\textbf{Scaled library:} The coefficients are fitted with columns scaled to unit root mean square over the observed states: $s_{m}:=\big(\frac{1}{N_{d}}\sum_{i=1}^{N_{d}}\Theta_{m}(y_{i})^{2}\big)^{1/2}$, $D:=\mathrm{diag}(s_{1},\ldots,s_{M})$, $\widetilde{\Theta}(x):=\Theta(x)D^{-1}$, and $\theta_{j}:=D\xi_{j}$, $j\in\{u,v\}$, so that $\widetilde{\Theta}(x)\theta_{j}=\Theta(x)\xi_{j}$; the model is unchanged. We write $\theta:=(\theta_{u},\theta_{v})\in\mathbb{R}^{20}$.
	
	\textbf{Difficulty:} In the quasi-steady-state regime, i.e., on the interval $[10,20]$, the $u$-data vary by about 0.3\% and the $v$-data by about 2\%  relative to their respective mean values. The observations thus cover only a small neighbourhood of the coexistence state $x^{*}$. Consequently, the monomial library matrix constructed from these observations is extremely ill-conditioned, with a smallest singular value of $2.3\times10^{-15}$ and a condition number of approximately $1.4 \times 10^{16}$. This creates three related challenges: (1) \textit{derivative estimation errors:} since $\dot{x} \approx 0$, small absolute errors can cause large relative errors; (2) \textit{insufficient excitation:} the data only cover a small neighbourhood of $x^{*}$ and so, the library terms hardly vary; (3) \textit{library collinearity:} candidate terms are nearly linearly dependent, allowing substantially different coefficient combinations to produce almost identical fits.
	
	\subsection{Physics-informed neural networks (PINNs) and sparse regression}\label{sec:2:1}
	
	\textbf{Networks:} Each state has its own network, $\hat{u}(\cdot;w_{u})$ and $\hat{v}(\cdot;w_{v})$, with weights $w = (w_{u}, w_{v})$ and $\hat{x}(t;w) := (\hat{u}(t;w_{u}), \hat{v}(t;w_{v})) \in \mathbb{R}^{2}$: four hidden layers of 64 neurons each, with $\tanh$ activation, the input time scaled to $[-1,1]$ with two Fourier features \cite{wang_eigenvector_2021}, $\sin(j\pi\tau)$ and $\cos(j\pi\tau)$ for $j = 1, 2$, and a softplus output. The derivative $\dot{\hat{x}}(t;w) = \partial \hat{x}(t;w)/\partial t$ is computed pointwise by automatic differentiation.
	
	\textbf{Loss function:} With collocation points $\bar{t}_{1},\ldots,\bar{t}_{N_{c}}$, $N_{c}=1,000$, on a grid over $[0,20]$, following the PINN framework of \cite{raissi_physics-informed_2019}, the loss is
	\begin{equation}\label{eqn:pinnloss}
		\mathcal{L}_{\mathrm{PINN}}(w,\theta):=\omega_{\mathrm{data}}\mathcal{L}_{\mathrm{data}}(w)+\omega_{\mathrm{ode}}\mathcal{L}_{\mathrm{ode}}(w,\theta)+\omega_{\mathrm{ic}}\mathcal{L}_{\mathrm{ic}}(w),
	\end{equation}
	with weights $(\omega_{\mathrm{data}},\omega_{\mathrm{ode}},\omega_{\mathrm{ic}})=(100,10,1)$, $\mathcal{L}_{\mathrm{data}}(w):=\frac{1}{N_{d}}\sum_{i=1}^{N_{d}}\|\hat{x}(t_{i};w)-y_{i}\|^{2}$, $\mathcal{L}_{\mathrm{ic}}(w):=\|\hat{x}(0;w)-x_{0}\|^{2}$ and
	\(\mathcal{L}_{\mathrm{ode}}(w,\theta):=\frac{1}{N_{c}}\sum_{l=1}^{N_{c}}\big\|\dot{\hat{x}}(\bar{t}_{l};w)-\tilde{f}\big(\hat{x}(\bar{t}_{l};w);\theta\big)\big\|^{2}\), where \(\tilde{f}(x;\theta):=\big(\widetilde{\Theta}(x)\theta_{u},\widetilde{\Theta}(x)\theta_{v}\big). \)
	
	\textbf{Regression at the collocation points:} For fixed networks, only $\mathcal{L}_{\mathrm{ode}}$ depends on $\theta$, and it is a linear least-squares problem: let $A\in\mathbb{R}^{N_{c}\times M}$, have rows $\widetilde{\Theta}(\hat{x}(\bar{t}_{l}))$, and let $b_{j}\in\mathbb{R}^{N_{c}}$ have entries $\dot{\hat{x}}_{j}(\bar{t}_{l})$, $j\in\{u,v\}$, then we have $\mathcal{L}_{\mathrm{ode}}=\frac{1}{N_{c}}\sum_{j\in\{u,v\}}\|A\theta_{j}-b_{j}\|^{2}$.
	
	\textbf{Recovery:} A mechanism is \textit{active} when its coefficient is nonzero. A \textit{support} $S$ is a set of active mechanisms, with at least one per equation, and $k := |S|$ is its size. There are $N_{k} = \binom{20}{k} - 2\binom{10}{k}$ supports of size $k$, e.g., $N_{7}=77,280$. For a support $S$, $c \in \mathbb{R}^{k}$ contains the coefficients of its active mechanisms, and $\Xi_{S}(c) \in \mathbb{R}^{M \times 2}$ has these entries and zeros elsewhere. A support with $a$ true and $b$ spurious mechanisms is written $a/b$; the true model is $7/0$, and only this support counts as recovered. We report the relative coefficient error separately.
	
	\section{Bayesian approach}
	\label{sec:3}
	
	In Bayesian model selection, competing models are scored by their \emph{evidence}, the marginal likelihood: the probability density of the data with the parameters integrated out against their prior~\cite{mackay_bayesian_1992,kass_bayes_1995}. The evidence contains an \emph{Occam factor} allowing models of different sizes to be compared on a common probabilistic scale. Sparse Bayesian learning~\cite{tipping_sparse_2001} maximises the evidence of a linear regression, and Bayesian versions of SINDy~\cite{zhang_robust_2018,fung_rapid_2025} apply it to regressions of derivatives. \textit{We compute the evidence at the level of the trajectories,} by refitting the candidate models to the observations, as for ranking ODE models in systems biology~\cite{vyshemirsky_bayesian_2008}; the networks only supply starting values.
	
	\textbf{The evidence of a support:} For a support $S$ with coefficients $c\in\mathbb{R}^{k}$, the candidate solution $x(\cdot;c)$ solves $\dot{x}(t;c)=f(x(t;c);\Xi_{S}(c))$ for $t>0$, with $x(0;c)=x_{0}$. Its values at $t_{1},\ldots,t_{N_{d}}$, stacked like $y$, form the prediction vector $X(c)\in\mathbb{R}^{n}$, and $r(c):=y-X(c)$ is the residual. 
	Although the observations $y$ are noise-free, we introduce an auxiliary Gaussian observation model $Y=X(c)+\varepsilon$, where $\varepsilon\sim\mathcal{N}(0,\sigma^{2}I_n)$. Here, $Y$ represents hypothetical observations, and $\sigma>0$ is a prescribed tolerance parameter rather than a measurement noise level. To represent uncertainty about the unknown coefficients, we assign a Gaussian prior $c \sim \mathcal{N}(0,\tau^{2}I_k)$, with a width of $\tau=2$ to cover coefficients of order one. The resulting \textit{evidence} is
	\[ Z(S)=\int_{\mathbb{R}^{k}}p(y \mid c,S) p(c \mid S) \mathrm{d}c=\dfrac{\int_{\mathbb{R}^{k}} e^{-\Phi(c)} \mathrm{d}c}{(2\pi\sigma^{2})^{\frac{n}{2}}(2\pi\tau^{2})^{\frac{k}{2}}}, \quad \Phi(c)=\frac{\|r(c)\|^{2}}{2\sigma^{2}}+\frac{\|c\|^{2}}{2\tau^{2}}. \]
	The \textit{maximum a posteriori} (MAP) coefficients
	\begin{equation}\label{eqn:MAPCoeffs}
		\hat{c} := \arg\min_{c\in\mathbb{R}^{k}}\Phi(c)
	\end{equation}
	are found by the Levenberg-Marquardt method \cite{marquardt_algorithm_1963}. It needs the sensitivity matrix $J(c):=\partial X(c)/\partial c\in\mathbb{R}^{n\times k}$, stacked from $\Lambda(t_{i})$, $i=1,\ldots,N_d$, where $\Lambda(t):=\partial x(t;c)/\partial c\in\mathbb{R}^{2\times k}$ solves the forward sensitivity equations $\dot{\Lambda}=\frac{\partial f}{\partial x}\Lambda+\frac{\partial f}{\partial c}$, $\Lambda(0)=0$; a candidate whose solution blows up is discarded. Expanding $\Phi$ to second order at $\hat{c}$ with $H:=\nabla^{2}\Phi(\hat{c})\approx J^{\top}J/\sigma^{2}+I_{k}/\tau^{2}$ gives $\int_{\mathbb{R}^{k}} e^{-\Phi} \mathrm{d}c \approx e^{-\Phi(\hat{c})}(2\pi)^{k/2}(\det H)^{-1/2}$ \cite{tierney_accurate_1986,mackay_bayesian_1992}, and the Laplace approximation of the log evidence is
	\begin{equation}\label{eqn:evidence}
		\log Z(S)\approx-\underbrace{\frac{\|r(\hat{c})\|^{2}}{2\sigma^{2}}}_{\text{fit}}-\underbrace{\frac{\|\hat{c}\|^{2}}{2\tau^{2}}}_{\text{prior}}-\underbrace{\Big(|S|\log\tau+\tfrac{1}{2}\log\det H\Big)}_{\text{Occam}}-\frac{n}{2}\log(2\pi\sigma^{2}).
	\end{equation}
	The last term is shared by all supports at one $\sigma$ and is dropped, so $\log Z$ at different $\sigma$ is not comparable. For two supports $S_1$ and $S_2$, the \textit{log Bayes factor} of $S_{1}$ over $S_{2}$ \cite{kass_bayes_1995} is $\log Z(S_{1})-\log Z(S_{2})$.
	
	The parameters $\sigma$ and $\tau$ jointly control the balance between data fit, coefficient regularization, and model complexity in the evidence. In particular, their ratio determines the strength of regularization and enters the Occam factor. Consequently, the choice of these parameters can affect the evidence and hence the selected model.
	
	\subsection{Description of our Bayesian approach}\label{sec:3:1}

	\begin{itemize}[
		label={},
		leftmargin=0.5em,
		itemindent=-0.5em,
		labelsep=3pt
		]
		\item \textbf{Stage 1:} Adam minimises \eqref{eqn:pinnloss} over $w$ and all of $\theta$ for 15,000 epochs.
		
		\item \textbf{Stage 2:} It consists of five steps:
		\begin{enumerate}
			\item {Sparse Bayesian Learning (SBL):} We fix the neural-network weights obtained in Stage 1 and re-estimate the ODE coefficients by applying SBL to $b_j \approx A\theta_j$, separately for each equation, using MacKay's fixed-point updates \cite{tipping_sparse_2001,mackay_bayesian_1992}. SBL suppresses some coefficients by setting them to zero, but all 20 candidate mechanisms remain in the model for subsequent optimisation.
			
			\item {$\ell_0$-IHT-HTP:} Starting from the SBL coefficients, we solve using iterative hard thresholding (IHT) \cite{blumensath_iterative_2009} with hard thresholding pursuit (HTP) \cite{foucart_hard_2011}, the optimisation $\min_{w,\theta}\mathcal{L}_{\mathrm{PINN}}(w,\theta)$ subject to $\|\theta\|_{\ell_0}\leq k$, for a given $k$ if known, or for $k=19,\ldots,2$ when $k$ is unknown, alternating between two steps:    
			\begin{itemize}[labelsep=0.33em]
				\item {$w$-step:} Fix $\theta$ and update the network weights $w$ (Adam, 200 epochs).
				\item {$\theta$-step:} Fix $w$ and perform up to 20 gradient steps on $\omega_{\mathrm{ode}}\mathcal{L}_{\mathrm{ode}}$ with step size $0.9/L$, where $L=2\omega_{\mathrm{ode}}\|A\|_2^2/N_c$ the Lipschitz constant of this gradient, each keeping the $k$ largest coefficients in magnitude (at least one per equation) and refitting them by HTP least squares on $A\theta_{j} \approx b_{j}$.
			\end{itemize}
			
			We perform at most 15 alternations, stopping if the support remains unchanged for three rounds. We then jointly optimise (Adam, 1,000 epochs) $w$ and $\theta$ with the support fixed.
			
			\item {Two-swap climb:} From the $\ell_{0}$-IHT-HTP support of the given $k$ (or for each $k$ when $k$ is unknown), every support within one or two swaps of mechanisms (at most 1,729 for $k=7$), is refitted through the ODE from least squares on the Stage~1 networks and scored using the Laplace approximation of the log evidence \eqref{eqn:evidence}: the climb moves to the best while the evidence rises, repeating until no further improvement is possible.
			
			\item Model selection: If $k$ is known, the support obtained from the two-swap climb is selected for the next step. If $k$ is unknown, the support with the highest log evidence among the final supports obtained for different values of $k$ is selected. This step selects only the support, not the final coefficients.   
			
			\item ODE coefficient refit: For the selected support, we estimate the coefficients using MAP optimisation \eqref{eqn:MAPCoeffs}. We run the optimisation twice, starting from least-squares estimates based on either the current neural networks or the Stage~1 networks. We retain the result with the higher log evidence and fix its coefficients $\hat{c}$ for the subsequent steps.
		\end{enumerate}
		
		\item \textbf{Stage 3:}  With the MAP coefficients $\hat{c}$ and support fixed, $\mathcal{L}_{\mathrm{PINN}}$ is minimised over $w$ (Adam, 5,000 epochs, then at most 300 L-BFGS steps).
	\end{itemize}
	
	Since the two-swap climb is a heuristic local search, it may fail to identify the highest-scoring support. To assess this possibility, we also perform an exhaustive search for comparison, replacing Step~3 of Stage~2. This search evaluates all admissible supports of size $k$ using the same coefficient-refitting procedure and Laplace-approximated log-evidence criterion.
	
	\subsection{Results of our Bayesian approach}
	\label{sec:3:2}
	
	All experiments, including the cases with known and unknown $k$ and the exhaustive search over $k=6,\ldots,10$, started from the same networks obtained in a single Stage~1 training run on one node of the RUC-HPC cluster.
	
	\textbf{Size known, $k=7$, $\sigma=2\times10^{-7}$:}
	Our Bayesian approach, incorporating the two-swap search, recovered the true $7/0$ support with a log evidence of $\log Z=-94.72$. The recovered model, rounded to two decimals, is given by
	\begin{equation*}
		\dot{u} = -0.20\, u + 1.10\, u^{2} - 0.20\, uv - 1.00\, u^{3}, \quad
		\dot{v} = 0.50\, u - 1.00\, v + 0.50\, uv,
	\end{equation*}
	with a relative coefficient error of $1.5\times10^{-6}$. The largest deviation, $2.4\times10^{-6}$, occurs in the coefficient of $u^{2}$, estimated as $1.1000023652$.
	
	\textbf{Size unknown, $\sigma=2\times10^{-7}$:}
	We applied our Bayesian approach at every size $k=2,\ldots,19$. The evidence selected $k=7$: same $7/0$ ($\log Z=-94.72$) support as with $k=7$ known, with the same relative coefficient error. Of the two-swap climbs at the other sizes, the best ended at $k=9$ ($-102.47$), with the seven true mechanisms and two spurious ones. Smaller models fitted the data poorly, while larger models were penalised for their additional complexity.
	
	\textbf{Exhaustive search,  $\sigma=2\times10^{-7}$:}
	To verify the evidence ranking, we exhaustively evaluated all 594,194 candidate supports of sizes $k=6,\ldots,10$. At $k=7$, each of the 77,280 supports was refitted from two initialisations, least squares on the Stage~1 networks and least squares based on the exact derivatives of the true solution, and both ranked the true model first. \textit{No support of sizes $6$ to $10$ beats the recovered true $7/0$ ($-94.72$) support}: the closest is a seven-term $2/5$ ($-97.34$) support and an eight-term $2/6$ ($-98.36$) support for any other size. Assuming equal prior probabilities over the 594,194 evaluated supports of sizes $k=6,\ldots,10$, the approximate posterior probability of the true model was $0.81$, conditional on the chosen $\sigma$ and $\tau$. The other $594,193$ share the remaining $0.19$.
	
\begin{figure}[tbp]
		\begin{minipage}[t]{0.50\linewidth}\vspace{0pt}
			\captionof{table}{The four best seven-term supports from the exhaustive search at $k=7$ at $\sigma=2 \times 10^{-7}$. $\log Z$: the evidence \eqref{eqn:evidence} of the refit; $\delta:=\|r(\hat{c})\|_2/\sqrt{n}$: misfit of the ODE refit; $\|\Delta x\|_{\infty}$: largest deviation from the true solution on $[0,20]$; $\mathcal{L}_{\mathrm{PINN}}$ after Stage~3, in $10^{-6}$.}
			\label{tab:top}
			\scriptsize\setlength{\tabcolsep}{5.0pt}
			\begin{tabular}{@{}lclll@{}}
				\hline\noalign{\smallskip}
				model & $\log Z$ & $\delta$ & $\|\Delta x\|_{\infty}$ & $\mathcal{L}_{\mathrm{PINN}}$\\
				\noalign{\smallskip}\svhline\noalign{\smallskip}
				$7/0$ & $-94.72$ & $2.9{\times}10^{-11}$ & $2.7{\times}10^{-8}$ & $2.377$\\
				$2/5$ & $-97.34$ & $4.3{\times}10^{-8}$ & $8.4{\times}10^{-2}$ & $2.297$\\
				$2/5$ & $-104.01$ & $6.8{\times}10^{-8}$ & $8.4{\times}10^{-2}$ & $2.225$\\
				$6/1$ & $-104.59$ & $5.8{\times}10^{-8}$ & $8.2{\times}10^{-4}$ & $2.324$\\
				\noalign{\smallskip}\hline
			\end{tabular}
		\end{minipage}\hfill
		\begin{minipage}[t]{0.47\linewidth}\vspace{0pt}
			\centering
			\includegraphics[width=0.94\linewidth]{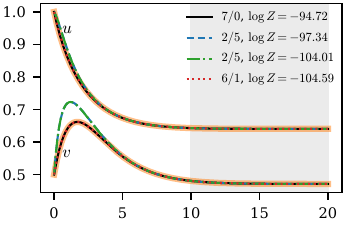}
			\caption{The four models of Table~\ref{tab:top} over the true solution (orange).}
			\label{fig:rivals}
		\end{minipage}
	\end{figure}

	Table~\ref{tab:top} and Fig.~\ref{fig:rivals} compare the true model with its three closest rivals at $k=7$. At fixed coefficients, the PINN losses of the four models differ by less than $7\%$, with the true model having the highest loss. Thus, a smaller PINN loss does not necessarily indicate the correct model.
	
	\textbf{Dependence on $\sigma$:}
	At $\sigma=10^{-5}$, a separate exhaustive search over all seven-term supports ranked the true model only 2,287th, while an incorrect $1/6$ support achieved higher Bayesian evidence. Thus, successful recovery of the true model depends on $\sigma$.
	
	\textbf{Comparison:} We compare our approach with five sparse model discovery methods. None of these recovers the exact model from the observed window and $x_{0}$ in any of the settings we implemented: SINDy with sequentially thresholded least squares on finite or smoothed finite differences \cite{Brunton_Proctor_Kutz_2016} (0 of 116), weak SINDy \cite{messenger_weak_2021} (0 of 174), ensemble SINDy \cite{fasel_ensemble-sindy_2022} (0 of 60), sparse Bayesian learning \cite{tipping_sparse_2001} (0 of 6) and sparse Bayesian regression \cite{hirsh_sparsifying_2022} (0 of 4): all run on PySINDy~\cite{kaptanoglu_pysindy_2022} except for sparse Bayesian learning regression, which PySINDy lacks, and so, we use our own implementation. Using exact derivatives of the true trajectory (a diagnostic only), standard SINDy with sequentially thresholded least squares succeeds in many of its settings: recovery is possible for this method under ideal conditions. 
	
	\textbf{Code and data availability:} A public version of the code is available at\\{\small\url{https://github.com/reymartsalcedo/ode-to-the-sparse-truth}}.

	\section*{Acknowledgements}
	RSL and CR acknowledge support through the Novo Nordisk Foundation grant NNF25OC0105537.

\end{document}